\documentclass[letterpaper, 10 pt, conference]{ieeeconf}  % Comment this line out if you need a4paper

\IEEEoverridecommandlockouts                              % This command is only needed if 
\usepackage{graphics} % for pdf, bitmapped graphics files
\usepackage{epsfig} % for postscript graphics files
\usepackage{mathptmx} % assumes new font selection scheme installed
\usepackage{times} % assumes new font selection scheme installed
\usepackage{amsmath} % assumes amsmath package installed
\usepackage{amssymb}  % assumes amsmath package installed

\usepackage{booktabs}

\usepackage{color}
\usepackage{mathrsfs}
\usepackage{algpseudocode}
\usepackage{multirow}
\usepackage{caption}
\usepackage{subcaption}
\usepackage{diagbox}
\usepackage{bm}

\usepackage[pagebackref=true,breaklinks=true,letterpaper=true,colorlinks,bookmarks=false]{hyperref}

\title{\LARGE \bf
Analyzing Infrastructure LiDAR Placement with Realistic LiDAR Simulation Library
}

\author{Xinyu Cai$^{*}$, Wentao Jiang$^{*}$, Runsheng Xu, Wenquan Zhao, Jiaqi Ma, Si Liu, Yikang Li$^{\dagger}$
\thanks{$^{*}$ Equally contributed to the work.}
\thanks{$^{\dagger}$ Corresponding author.}
\thanks{Xinyu Cai, Wenquan Zhao and Yikang Li are with Autonomous Driving Group, Shanghai AI Laboratory, China. {\tt\small \{caixinyu, zhaowenquan, liyikang\}@pjlab.org.cn}
}
\thanks{Wentao Jiang, and Si Liu are with Institute of Artificial Intelligence, Beihang University. {\tt\small \{jiangwentao, liusi\}@buaa.edu.cn}
}
\thanks{Runsheng Xu and Jiaqi Ma are with University of California, Los Angeles, USA.{\tt\small \{rxx3386, jiaqima\}@ucla.edu} }
}

\begin{document}

\maketitle
\thispagestyle{empty}
\pagestyle{empty}

%%%%%%%%%%%%%%%%%%%%%%%%%%%%%%%%%%%%%%%%%%%%%%%%%%%%%%%%%%%%%%%%%%%%%%%%%%%%%%%%

\begin{abstract}

Recently, Vehicle-to-Everything~(V2X) cooperative perception has attracted increasing attention.
Infrastructure sensors play a critical role in this research field, however, how to find the optimal placement of infrastructure sensors is rarely studied.
In this paper, we investigate the problem of infrastructure sensor placement and propose a pipeline that can efficiently and effectively find optimal installation positions for infrastructure sensors in a realistic simulated environment.
To better simulate and evaluate LiDAR placement, we establish a Realistic LiDAR Simulation library that can simulate the unique characteristics of different popular LiDARs and produce high-fidelity LiDAR point clouds in the CARLA simulator.
Through simulating point cloud data in different LiDAR placements, we can evaluate the perception accuracy of these placements using multiple detection models.
Then, we analyze the correlation between the point cloud distribution and perception accuracy by calculating the density and uniformity of regions of interest.
Experiments show that the placement of infrastructure LiDAR can heavily affect the accuracy of perception.
We also analyze the correlation between perception performance in the region of interest and LiDAR point cloud distribution and validate that density and uniformity can be indicators of performance.
Both the RLS Library and related code will be released at \url{https://github.com/PJLab-ADG/PCSim}.

\end{abstract}

%%%%%%%%%%%%%%%%%%%%%%%%%%%%%%%%%%%%%%%%%%%%%%%%%%%%%%%%%%%%%%%%%%%%%%%%%%%%%%%%

\section{Introduction}

Perceiving the driving environment precisely is important for autonomous driving.
With recent advancements in deep learning, the robustness of single-vehicle perception algorithms has demonstrated significant improvement in several tasks such as object detection \cite{lang2019pointpillars,2018Complex,2017Vote3Deep,2018YOLO3D,2017PointNet,2014Beyond,2017VoxelNet}.
Compared to single-vehicle perception, utilizing both vehicle and infrastructure sensors brings many significant advantages, including providing a global perspective far beyond the current horizon and covering blind spots.
Current advances in Vehicle-to-Everything~(V2X) communications have made it possible to utilize data from infrastructure sensors \cite{xu2022v2x,bhover2017v2x,olaverri2018connection}.

Recently, some literature has considered the LiDAR perception problem from the viewpoint of LiDAR placement \cite{chen2021pole,hu2022investigating,ma2021perception}, which is a new perspective and crucial since improper LiDAR placements may cause poor-quality sensing data and thus lead to inferior performance of perception.
With the rapid development of V2X applications, knowing how to choose the optimal placement to maximize the benefits brought by infrastructure sensors is necessary.
However, existing research on LiDAR placement mainly focuses on vehicles and ignores the placement problem for infrastructure sensors.
The placement of infrastructure LiDAR has a higher degree of freedom -- both the $xyz$ positions and roll, pitch, and yaw angles need to be considered.
Therefore, in this paper, we aim to analyze the interplay between infrastructure LiDAR sensor placement and perception performance.

\begin{figure}[!t]
   \centering
   \includegraphics[width=1.0\linewidth]{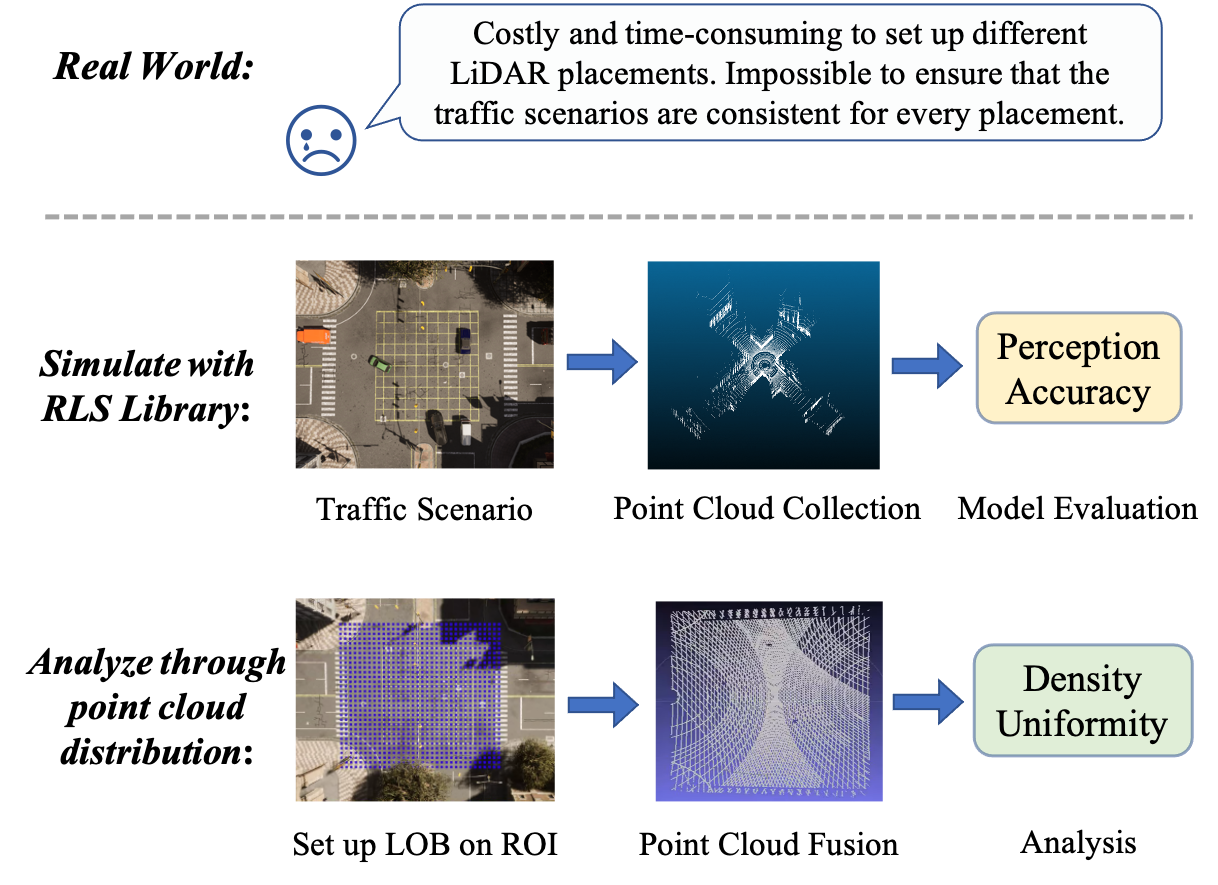}
   % \vspace{-1mm}
   \caption{
       It is infeasible to evaluate infrastructure LiDAR placements in the real world. Thus, we propose to simulate and then evaluate the placements using the RLS library. We also analyze the correlation between point cloud distribution and perception performance.
   }
%   \vspace{-2mm}
   \label{fig1}
\end{figure}

The most straightforward way to find the optimal installation position is to evaluate the perception performance of diverse LiDAR placement choices in the real world.
Nevertheless, such an approach is infeasible, as it is costly and time-consuming.
Furthermore, it is impossible to ensure the traffic scenarios are identical when evaluating different placements, which will bring unfair comparisons, as shown in Figure \ref{fig1}.
One alternative solution is to use a high-fidelity simulator to reproduce the real-world environment consistently and assess the placement strategies with low costs.
Existing works \cite{hu2022investigating,ma2021perception,manivasagam2020LiDARsim} choose the popular CARLA \cite{dosovitskiy2017carla} simulator for data collection and evaluation.
But this simulator only supports Surround LiDAR with uniform beams and cannot simulate the realistic physical characteristics of different LiDAR sensors, which restricts the generalization to the real-world environment.

To obtain accurate and realistic LiDAR points clouds and facilitate V2X applications in the real-world environment, we develop the Realistic LiDAR Simulation (RLS) library, which precisely simulates the physical characteristics of different LiDARs and can be installed as a plugin in CARLA.
Our RLS library contains 14 popular LiDAR devices of 3 types, including Surround LiDAR, Solid State LiDAR, and Risley Prism LiDAR. We borrow these LiDARs and conduct evaluation and analysis on realistic point cloud data produced by these LiDARs.
Besides the characteristics of different LiDARs, our RLS library can also simulate different LiDAR beams, motion distortions, and ghosting effects, which are frequently seen in real point cloud data.
With the RLS library, we can fix the traffic scenario to obtain realistic point cloud data for each LiDAR placement, and evaluate the perception accuracy for each placement.

However, the simulate-and-evaluate pipeline is quite effort-costly, which includes the process of LiDAR deployment, data collection, model training, and performance evaluation.
This pipeline also depends on the training and testing traffic scenario.
% When solving the infrastructure LiDAR placement problem, we find that traffic scenario perception performance and point cloud distribution are related.
To facilitate the design of LiDAR placement, we dive into the correlation between perception performance and point cloud distribution.
We leverage the density and uniformity of point cloud data as two surrogate metrics and analyze how they affect detection accuracy.
The surrogate metrics are helpful for quick evaluation of placement and help to find better placement for infrastructure LiDARs.

Our contributions are summarized as follows:
\begin{itemize}
    \item To the best of our knowledge, we are the first to study infrastructure LiDAR placement problems. We propose a simulation library and toolkit to evaluate different infrastructure LiDAR placements, which facilitates its applications in the real-world environment.
    \item We propose the Realistic LiDAR Simulation (RLS) library that can simulate 14 popular LiDAR devices with their unique characteristics, including LiDAR beam, motion distortion, and ghosting object effect. With the RLS library, we can evaluate different placements in a realistic simulated environment.
    \item We evaluate multiple LiDAR placements and investigate the relationship between the accuary of perception and the distribution of point cloud data. Through our observation, the density and uniformity of point cloud data can help predict the accuracy of perception, which are helpful for quick evaluation.
\end{itemize}

\begin{figure*}[!t]
   \centering
   \includegraphics[width=0.9\linewidth]{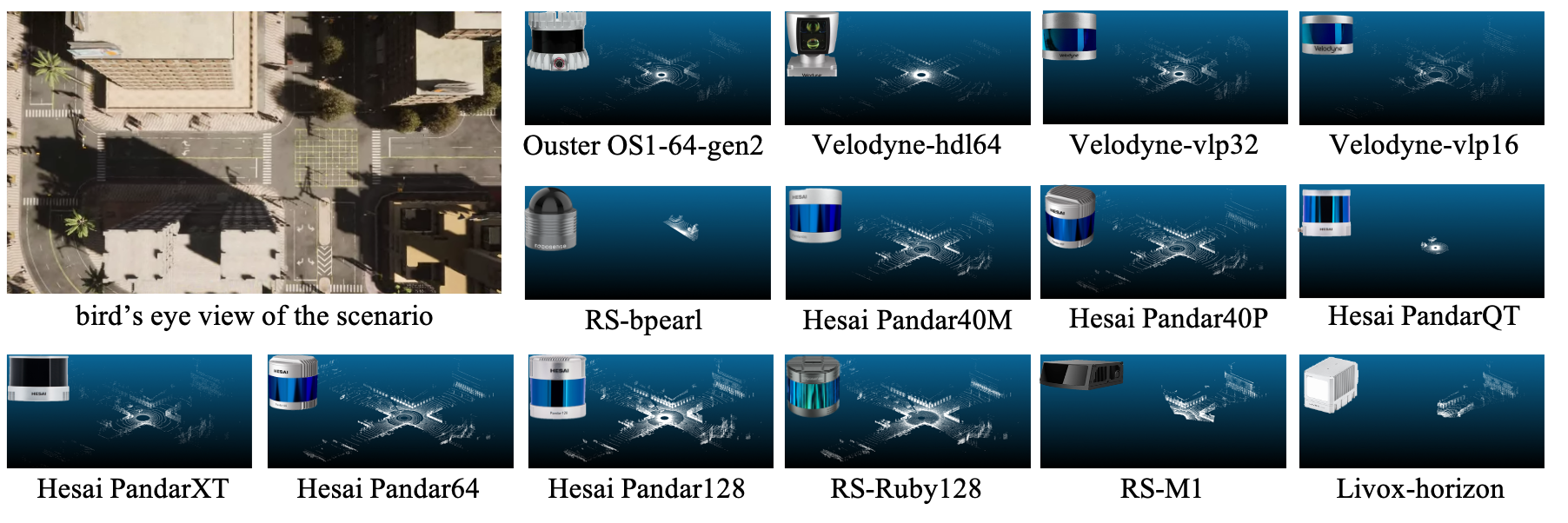}
%   \vspace{-1mm}
   \caption{
        With the newly proposed RLS library, we can simulate realistic point cloud data using 14 popular LiDAR devices of 3 types, including Surround LiDAR, Solid State LiDAR, and Risley Prism LiDAR.
        The RLS library can restore the unique characteristics of different kinds of LiDAR.
   }
%   \vspace{-3mm}
   \label{LiDARsim}
\end{figure*}

\section{Related Work}

\textbf{LiDAR simulation.}
Previous research work related to LiDAR simulation focused more on target modeling within the LiDAR perception range, real-world simulation reconstruction, and point cloud rendering generation. Manivasagam et al. \cite{manivasagam2020LiDARsim} used real-world data to reconstruct the surrounding traffic scene and obtained the LiDAR point cloud data of the transformed traffic scenario by simulation. Caccia et al. \cite{caccia2019deep} used a deep generative model to reconstruct high-quality LiDAR point cloud data. Hahner et al. \cite{hahner2019semantic, hahner2021fog, hahner2022lidar} studied the LiDAR point cloud simulation work in adverse weather. They investigated the point cloud characteristics of bad weather and used the real LiDAR point cloud as the input to simulate the point cloud data in adverse weather such as foggy and snowy days. But their work ignores the differences between different technology routes and different models of LiDARs. At the same time, their work did not involve the simulation of LiDAR physical characteristics.

\textbf{V2X/multi-agent perception.}
V2X perception aims to detect objects in traffic environments with sensors on vehicles and other devices, which is also a kind of multi-agent sensor fusion problem.
Current research on V2X perception mainly focuses on V2V (Vehicle-to-Vehicle) and V2I (Vehicle-to-Infrastructure).
V2X perception studies how to fuse visual information from neighboring vehicles and infrastructure sensors to promote perception performance.
V2X perception can be broadly categorized into early, intermediate, and late fusion.
Early fusion \cite{gao2018object} directly transformed the raw data and fused them to form a holistic view for perception.
Intermediate fusion \cite{li2021learning,chen2017multi,wang2020v2vnet,xu2022opv2v,xu2022v2x, lei2022latency, su2023uncertainty} extract the intermediate neural features from on each agent’s observation and then fused for perception.
While late fusion \cite{melotti2020multimodal, fu2020depth, caltagirone2019lidar, 2012Car2X}  fuses the perception results and uses NMS to produce final results.
Though early fusion provides valuable scenario context which late fusion can not provide, it requires large transmission bandwidth.
Therefore, intermediate fusion has attracted research's attention since its good balance between accuracy and transmission bandwidth.
Several intermediate fusion methods have been proposed for V2X perception recently.
\cite{chen2019f} aggregates the features from other agents with equal weights. 
V2VNet \cite{wang2020v2vnet} proposes a spatial-aware message passing mechanism to jointly reason detection and prediction.
\cite{2020Learning} regresses vehicle localization errors with consistent pose constraints to attenuate outlier messages.
DiscoNet \cite{li2021learning} leverages knowledge distillation to enhance training by constraining the corresponding features to the ones from the network for early fusion. However, intermediate fusion for V2X is still in its infancy.
Recently, V2X-ViT \cite{xu2022v2x} presented a unified transformer architecture for V2X perception, which can capture the heterogeneous nature of V2X systems with strong robustness against various noises.
In this paper, we study the influence of infrastructure LiDAR placement on the V2X/multi-agent accuracy using the aforementioned state-of-the-art methods.

% \textbf{LiDAR placement for autonomous driving.}
\textbf{LiDAR placement.}
Previous research on LiDAR placement mainly focused on the optimization placement of Surround LiDAR for autonomous driving.
To tackle the surrounding LiDAR placement problem for autonomous driving, Ma et al. \cite{ma2021perception} proposed a new method based on Bayesian theory conditional entropy to evaluate vehicle sensor configuration.
Liu et al. \cite{liu2019should} proposed a bionic index - volume to the surface ratio (VSR) to achieve the optimal LiDAR configuration problem of maximizing utility.
Dybedal et al. \cite{dybedal2017optimal} used the mixed integer linear programming method to solve the problem of finding the best placement position of 3D sensors.
Hu et al. \cite{hu2022investigating} proposed a method for optimizing the placement of vehicle-borne LiDAR and compared the perceptual performance effects of many traditional schemes.
Kim et al. \cite{kim2019placement} studied how to increase the number of point clouds in a specific area and reduce the dead zone as much as possible.
However, existing works rarely consider the impact of different scanning technology routes of LiDAR on the placement problem. They only consider the placement problem of Surround LiDAR, while ignoring the placement problem of Solid State LiDAR, and Risley Prism LiDAR. Moreover, most of the LiDAR placement methods for the vehicle only consider the translation of the position and rarely consider the rotation, in other words, the transformation of the Euler angle. Because changing the placement position is more costly for infrastructure LiDAR placement problems, rotation transformation is crucial to the placement of infrastructure LiDARs. Therefore, the LiDAR placement methods for vehicles are often not applicable to Infrastructure LiDAR placement problems. Using the original LiDAR physical model of the simulation framework means that the gap between the scanning models, physical characteristics such as LiDAR motion distortion, and ghosting object effect between the real LiDAR and the LiDAR that comes with the simulation library are ignored, which will affect the persuasiveness of the final conclusion in the real environment.

\section{Realistic LiDAR Simulation Library}

Our RLS library simulates 14 popular LiDAR of 3 types, including Surround LiDAR, Solid State LiDAR, and Risley Prism LiDAR.
To obtain the physical characteristics of these LiDARs, we borrow these LiDARs and conduct evaluation and analysis on realistic point cloud data.
The characteristics we evaluate include the maximum detection range, blind area range, ranging accuracy, angular resolution, noise detection, etc.
After obtaining and summarizing the physical characteristics of these LiDARs, we build the RLS library based on the CARLA simulation framework.
Our RLS library mainly includes three new features compared to existing LiDAR simulation works, including LiDAR beam simulation, motion distortion simulation, and virtual shadow simulation.
More detail about the RLS library can be found in the link in the abstract.

\textbf{LiDAR beam simulation.}
Existing open source and paid simulation frameworks \cite{dosovitskiy2017carla,manivasagam2020LiDARsim} are able to support Surround LiDAR beam simulation of the uniform beam. The optional attributes often include channels, range, points per second, rotation frequency, upper \& lower FOV, etc.
However, existing works can not support the simulation of LiDAR beam that scans according to other beam modes, such as non-uniform beam Surround LiDAR, Solid State LiDAR, and Risley Prism LiDAR.  
% While retaining the original pipeline of LiDAR point cloud generation in the simulation framework,
To achieve beam simulation, we propose a point cloud generation pipeline for the LiDAR of the aforementioned three technical routes in our newly proposed RLS library.
1) For non-uniform beam Surround LiDAR, we calculate the altitude angle of each laser emitting beam according to the angular resolution distribution in different regions.
2) For Solid State LiDAR, we use the Lissajous equation as the base function to simulate the periodic change of the scanning beam direction caused by the vibration of the transverse and longitudinal axes of the MEMS galvanometer.
3) For the Risley Prism LiDAR, we use the officially released point cloud explanatory notes to obtain the azimuth angle and altitude angle point by point and complete the point-by-point simulation in the CARLA simulator.

\begin{figure}[!t]
   \centering
   \includegraphics[width=1.0\linewidth]{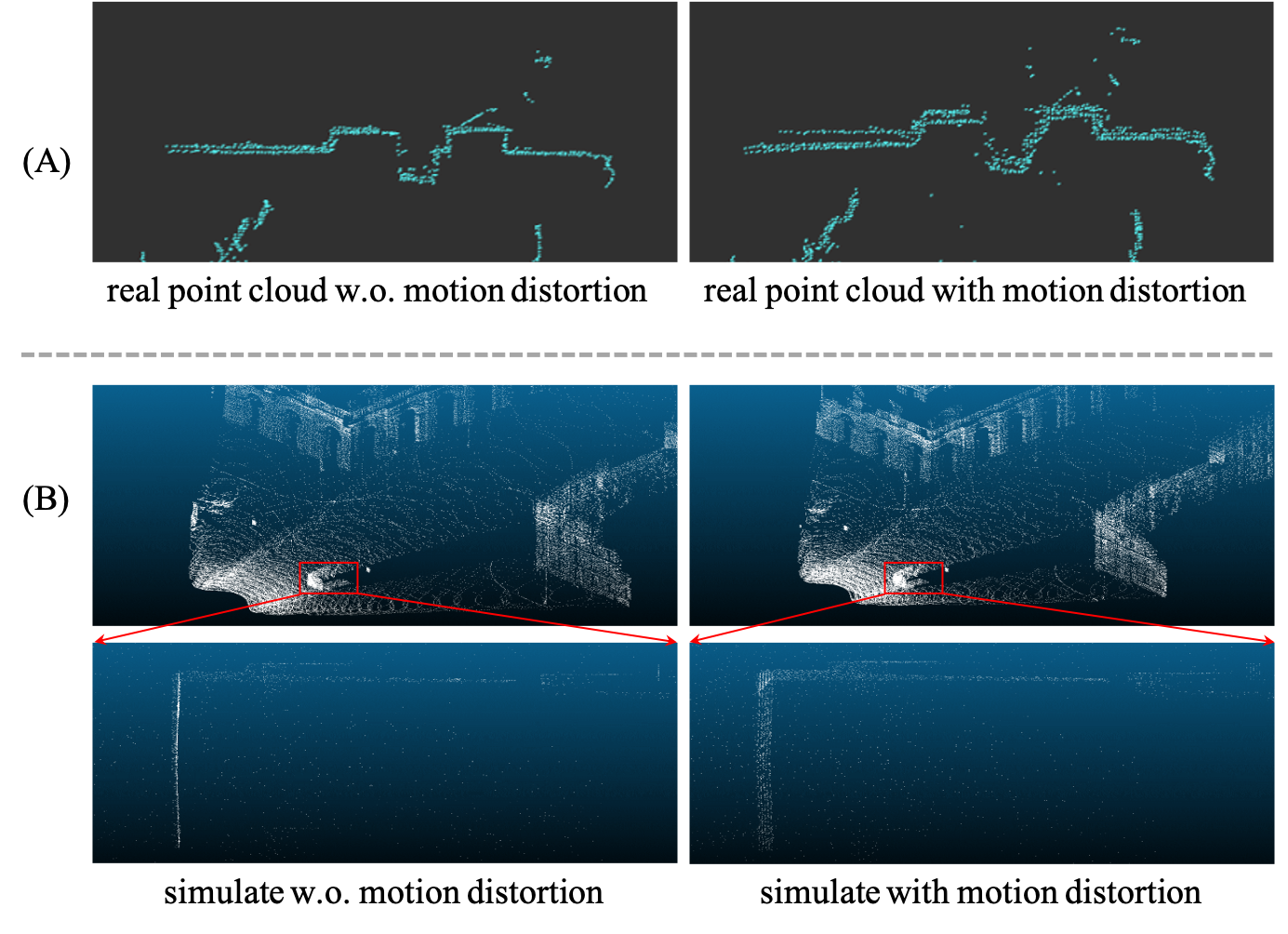}
   % \vspace{-1mm}
   \caption{
        (A) Motion distortion in real point cloud data. (B) Motion distortion simulated by RLS library.
       Our RLS library simulates the motion distortion effect, which brings realistic point cloud data for moving objects.
   }
  \vspace{-3mm}
   \label{motion}
\end{figure}

\begin{figure}[!t]
   \centering
   \includegraphics[width=1.0\linewidth]{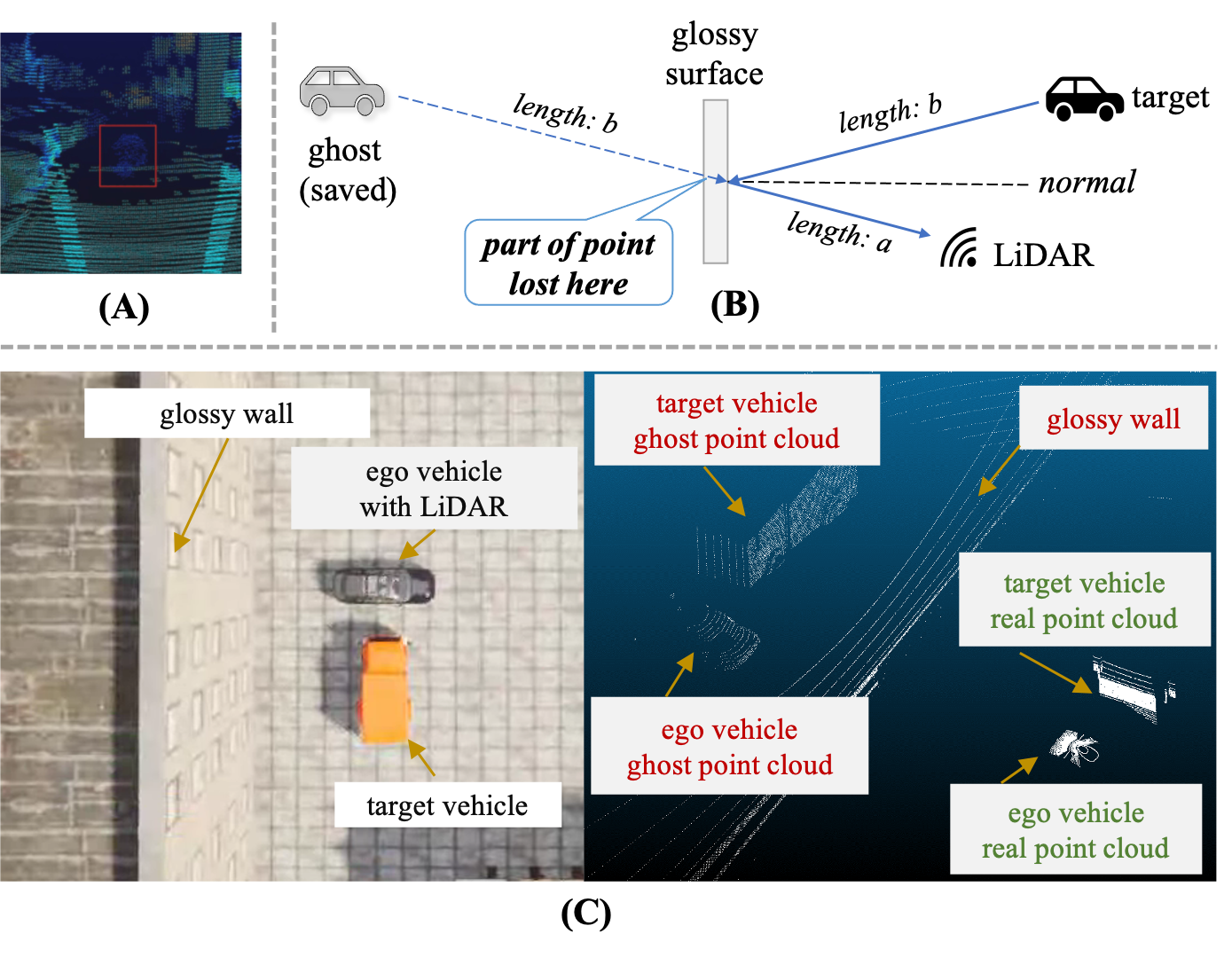}
   % \vspace{-1mm}
   \caption{
       (A) Ghosting object effect in real point cloud data. (B) A kind of cause of ghosting object effects. (C) A scene simulated by our RLS library.
   }
  \vspace{-3mm}
   \label{ghost}
\end{figure}

\textbf{Motion distortion simulation.}
In a real traffic scenario, both the traffic participants and the LiDAR are constantly moving during the period of each data frame collected by the LiDAR.
When the relative position between LiDAR and the surrounding environment changes, whether it is dithering or moving, the collected point cloud data will have motion distortion.
If the movement speed is fast enough, the motion distortion can not be ignored. As shown in Figure \ref{motion} (A), the point cloud of the wall is thickened due to LiDAR motion distortion.
In most other simulation frameworks, considering the simulation performance, the default approach is that the point cloud acquisition of each frame of LiDAR is only triggered in one simulation frame. At this time, the surrounding environment is completely still, and "snapshot" point cloud data is obtained instead of continuously collecting point cloud data.
In the LiDAR simulation library we developed, we set a switch for the LiDAR motion distortion simulation function.
If the function is switched on, to simulate continuous point cloud acquisition, the system frequency of the simulation frame will increase to $N$ times the default frequency, and the number of LiDAR acquisition points corresponding to each system simulation frame will decrease to $\frac{1}{n}$.
In order to keep the LiDAR acquisition frequency and points consistent with the original one, the LiDAR simulation point cloud collected by each system frame will be temporarily stored in the cache until the $N$-frame system simulation is completed.
Then the cached data will be uniformly output as a frame of point cloud data.
An example of our simulation is shown in Figure \ref{motion} (B), the point cloud of the vehicle is thickened due to LiDAR motion distortion.

\textbf{Ghosting object effect simulation.} 
In LiDAR point cloud data, ghosting object effects have multiple causes. One of the ghosting object effects in real point cloud data is shown in Figure \ref{ghost} (A). A considerable part of the effect is caused by the following reason. When the LiDAR beam hits a surface with a high degree of smoothness and high reflectivity, the specular reflection will be triggered. 
The reflected light will re-hit other objects close to the glossy mirror. The echo light will then return to the LiDAR receiving end along the transmitting optical path, as shown in Figure \ref{ghost} (B).
However, most LiDARs cannot tell whether the outgoing light is specularly reflected, and only considers that there is a target object at the distance of $a+b$ in the straight line direction of the current optical path, and store the point cloud data of this part.
In the real world, glass, water, etc. are easy to trigger the ghosting object effect.
In order to improve the consistency between simulation and the real world, it is necessary to simulate the ghosting object effect of LiDAR.

We simulate the ghosting object effect as follows:
First, we establish a target-reflector smoothness library for different target types that can be hit by LiDAR in the simulation environment.
We calculate the unit vector of the direction of the outgoing light through the azimuth angle and the altitude angle of the outgoing light of the LiDAR.
Then, the semantic information of the hit object is obtained through the simulation framework, including the target type hit by the outgoing light of the LiDAR, the direction vector of the incident angle, and the normal direction vector of the reflecting surface.
When the smoothness of the reflective surface of the hit object exceeds the threshold (such as water surface, or glass), the phantom feature is triggered, and this part of the point cloud is removed, which will not be included in the point cloud data.
After that, the reflected light direction information is calculated by the unit vector of the outgoing light, the normal vector of the hitting surface, the size of the incident angle, and the coordinate information of the first hitting point.
Finally, the reflected light detects the target near the surface. 
We replace part of the point cloud at the glossy surface with mirror-symmetric ghost point cloud data, to complete the ghost effect simulation.
% and use the mirror-symmetrical data of the point cloud data that hits the target for the second time to replace the point cloud at the mirror surface, thus completing the ghosting object effect simulation. 
As shown in Figure \ref{ghost} (C), we set the ratio of triggering ghosts in all beams of the LiDAR to 50\%, which indicates 50\% of the points that hit the glossy wall generate the point cloud of the wall, and the other points generate the ghost point cloud.

% The algorithm flow and effect diagram of virtual shadow simulation are shown in the figure below.

% \section{Methodology}
\section{Analyzing Infrastructure LiDAR Placement}

\subsection{Simulate-and-Evaluate with RLS Library}

With the newly proposed RLS Library, we are able to simulate realistic point cloud data produced from different kinds of LiDARs.
Given a specific infrastructure LiDAR placement (like setting the LiDAR on two-way traffic lights), a straightforward way to evaluate its perception performance is to simulate the traffic scenario and evaluate the precision of object detection.
For fair comparisons, we use the same training and testing traffic scenario (same scenes, same cars, same trajectories), for every infrastructure LiDAR placement, and evaluate them with the same V2X/multi-agent perception models.
A higher average perception of 3D object detection indicates better infrastructure LiDAR placement.

% \subsection{Surrogate Metric: Placement Score}
\subsection{Analysis using Point Cloud Distribution}

As mentioned above, the pipeline of simulate-and-evaluate, which includes data collection, model training, and performance evaluation is still effort-costly.
Although we use the same traffic scenario for experiments, the evaluation still depends on the training and testing traffic scenario, i.e., training and testing data.
To accelerate the evaluation of infrastructure LiDAR placement and to avoid the influence of traffic scenarios, we design two traffic-irrelevant surrogate metrics, Infrastructure Density, and Infrastructure Normalized Uniformity Coefficient.
These two metrics help us evaluate the placement of LiDARs quickly for choosing better placement plans.

\begin{figure}[!t]
   \centering
   \includegraphics[width=1.0\linewidth]{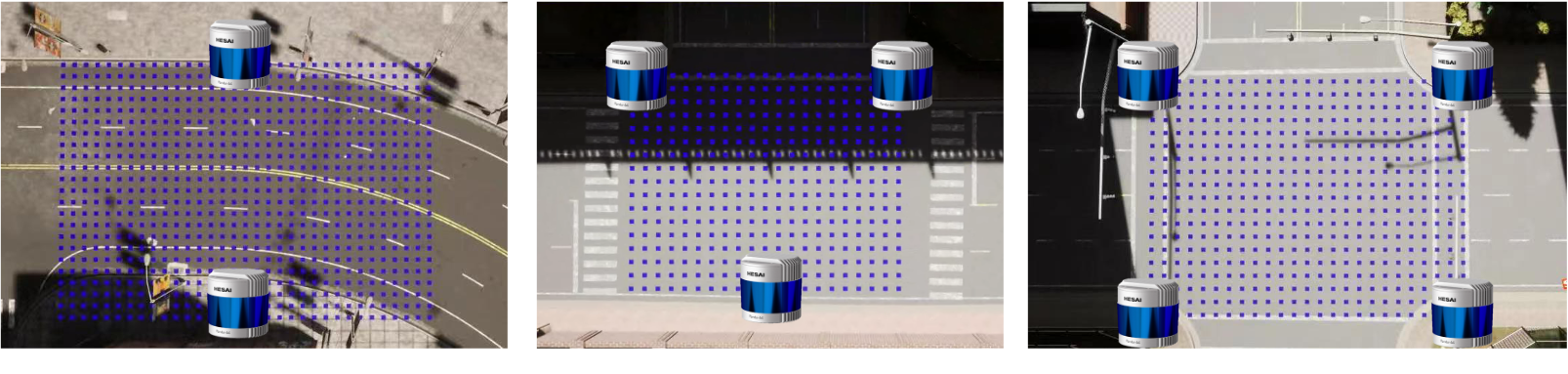}
   % \vspace{-1mm}
   \caption{
       The InfraLOB we set in three simulation scenes. InfraLOB is a virtual region of interest, where we expect infrastructure sensors to perceive the environment precisely.
   }
%   \vspace{-2mm}
   \label{lob}
\end{figure}

\textbf{Infrastructure LOB in ROI.}
LiDAR Occupation Board (LOB) \cite{kini2020sensor} is originally proposed for evaluating LiDAR on vehicles, where LiDAR Occupation Boards are several real boards that are placed around the vehicle LiDAR to evaluate the number of effective LiDAR points.
Inspired by this idea, we design a virtual LOB used for infrastructure LiDAR placement problems, named InfraLOB.
The difference between LOB and InfraLOB is that InfraLOB is a virtual region of interest, where we expect infrastructure sensors to perceive the environment precisely.
According to the actual situation of the intersection, taking into account the possible installation location of LiDAR and the traffic flow at the intersection, we delineate a rectangular region as large as possible as our ROI. This virtual region is our InfraLOB, as shown in Figure \ref{lob}.

\textbf{Infrastructure Density and Infrastructure Normalized Uniformity Coefficient.}
To quantify point cloud distribution in InfraLOB, we use Infrastructure Density(InfraD) and Infrastructure Normalized Uniformity Coefficient(InfraNUC).
InfraD describes the density of point cloud within the InfraLOB, which can be calculated as,
\begin{equation}
    \begin{aligned}
        InfraD &= \frac{N}{S}
    \end{aligned}
\end{equation}
where $N$ is the number of points within the InfraLOB region, $S$ is the area of InfraLOB region.

InfraNUC is a variant of Normalized Uniformity Coefficient (NUC) \cite{li2019pu}. NUC is a measure of the overall uniformity of the point set across all objects in the test dataset.
In our problem, there is only one object, the InfraLOB described above. We propose InfraNUC to quantify the uniformity of InfraLOB point cloud distribution.
First, according to the size of the region of interest, calculate the area of InfraLOB. Then, within the area, the semantic information of the LiDAR is used to obtain the label of the target road surface, and the corresponding point cloud data in the InfraLOB is filtered out. We then randomly select a certain number of equally sized disks in the InfraLOB region. Traverse all disks and calculate the number of point clouds in each disk and their average. Then calculate the standard deviation of the points in the disk.
InfraNUC can be calculated as,
\begin{equation}
    \begin{aligned}
        avg &= \frac{1}{D} \sum_{i=1}^D \frac{n_i}{N \cdot p} \\
        InfraNUC &= \sqrt{\frac{1}{D} \sum_{i=1}^D\left(\frac{n_i}{N \cdot p}-avg\right)}
    \end{aligned}
\end{equation}
where $n_i$ is the number of points within the disk region, $N$ is the number of points within the InfraLOB region, $D$ is the number of disks, $p$ is the ratio of disk area to the InfraLOB area.
$avg$ is the average of the number of point clouds in all disks.
$InfraNUC$ is the normalized uniformity coefficient, which quantifies the point cloud distribution uniformity within the InfraLOB region.

\section{Experiments}

In this section, we aim to analyze how infrastructure LiDAR placement influences perception performance using our newly proposed Realistic LiDAR Simulation library.

\subsection{Experimental Setup}

\textbf{Experiment environment}.
We leverage the realistic self-driving simulator, CARLA \cite{dosovitskiy2017carla}, in our experiments.
Due to the requirement of fairly evaluating different LiDAR placements with all other environmental factors fixed, we can use CARLA to simulate realistic scenarios which have the same vehicle’s trajectories but with different infrastructure LiDAR placement.
We build our training and testing dataset upon the simulated scenarios of OPV2V \cite{xu2022opv2v} dataset.
Specially, we leverage the defined vehicle’s trajectories in the scenarios of OPV2V and collect the point cloud data using the proposed RLS library in the CARLA simulator.
For the infrastructure LiDAR placement experiments, we use the 64-channel Hesai Pandar64 LiDAR in the RLS library and collect the point cloud data at the frequency of 10 Hz in CARLA.

\textbf{Benchmark models.}
We leverage multiple V2X/multi-agent perception models for evaluation including intermediate fusion, late fusion, which gathers all detected outputs from agents and applies Non-maximum suppression to produce the final results, and early fusion, which directly aggregates raw LiDAR point clouds from nearby sensors.
For intermediate fusion strategy, we evaluate using four state-of-the-art approaches including OPV2V \cite{xu2022opv2v},  F-Cooper \cite{chen2019f}, V2VNet \cite{wang2020v2vnet}, and V2X-ViT \cite{xu2022v2x}.
All the models use PointPillar \cite{lang2019pointpillars} as the backbone.
We use benchmark models to evaluate the point cloud data collected in different placements of LiDARs. The mean of APs is regarded as the final perception performance.

\begin{figure}[!t]
   \centering
   \includegraphics[width=0.9\linewidth]{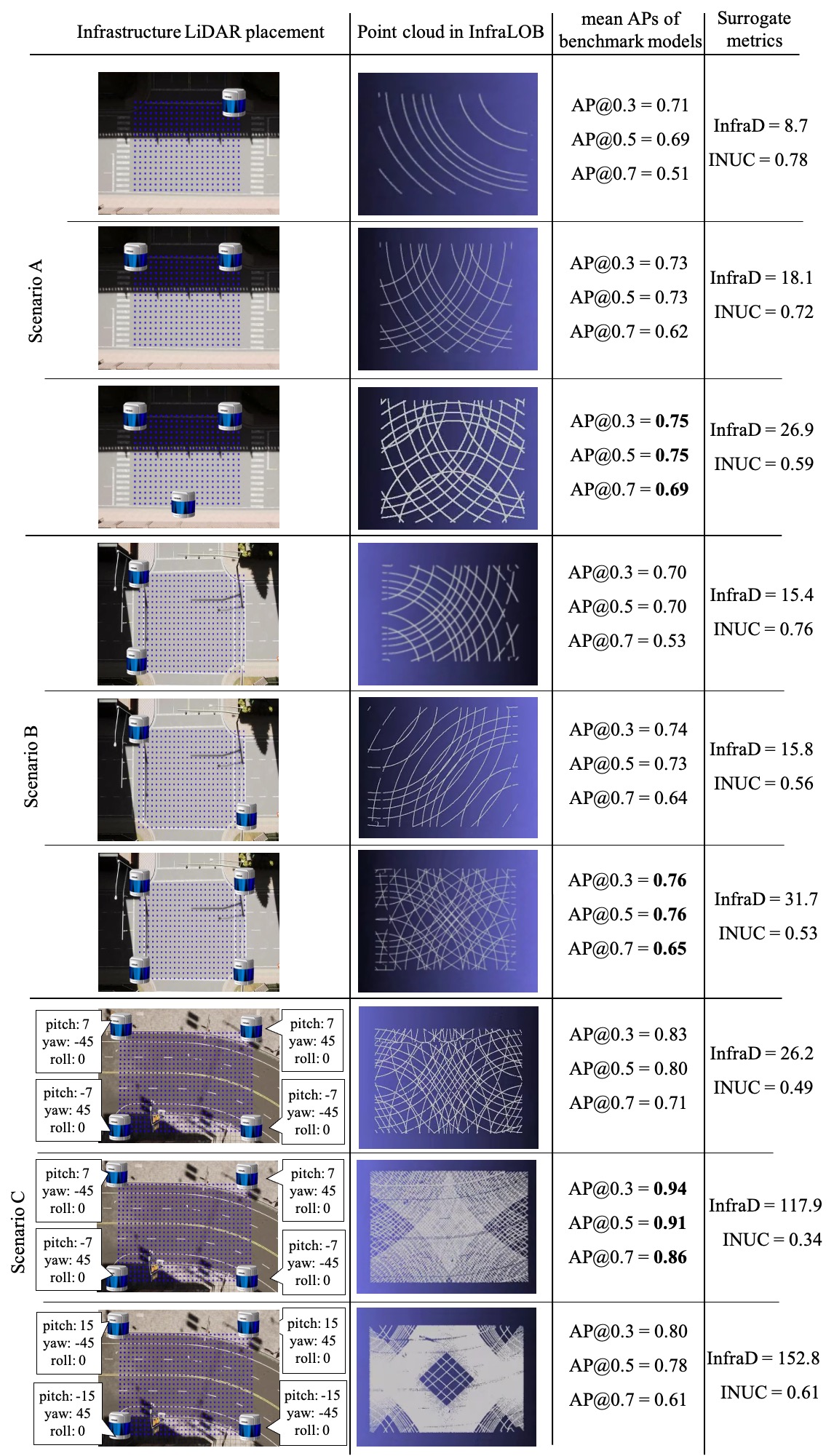}
   % \vspace{-1mm}
   \caption{
       We visualize 9 Infrastructure LiDAR placements in 3 different scenarios and evaluate the mean of accuracy. We also calculate the density and uniformity of the region of interest and analyze the correlation between accuracy and these metrics. Note that INUC indicates InfraNUC.
   }
%   \vspace{-2mm}
   \label{placements}
\end{figure}

\subsection{Experiments for Infrastructure LiDAR Placement}

% \noindent\textbf{Relationship between perception accuracy and point cloud distribution.}
We conduct experiments to evaluate different infrastructure LiDAR placements.
First, we use the simulate-and-evaluate pipeline to obtain the mean of average precision using the aforementioned benchmark models in testing scenarios.
Then, we try to investigate the relationship between perception accuracy and point cloud distribution in the region of interest under different infrastructure LiDAR placement plans.
To calculate the InfraNUC and InfraD of different LiDAR placements, we remove the traffic flow and keep infrastructure LiDARs unchanged to collect point cloud data.
We delimit the InfraLOB areas on the region of interest manually, which are the intersection area in our experiments.
After that, we segment the point cloud data and objects in the InfraLOB area and calculate InfraD and InfraNUC for every InfraLOB in testing scenarios.

Here, we show three common traffic scenarios in the test set as examples for evaluating infrastructure LiDAR placements, as shown in Figure \ref{placements}.
To visualize the point cloud distribution in InfraLOB, we project the point cloud data into the world coordinate system.
Scenario A is the T-junction of urban roads.
We place one, two, and three LiDARs at the T-junction to analyze how the number of LiDARs will affect the perception accuracy.
It can be observed that more LiDAR brings higher InfraD and lower InfraNUC, which also leads to performance gain.
Scenario B is the intersection of urban roads.
We first place 2 LiDARs at different locations as shown in rows 5 and 6 in Figure \ref{placements}.
Although these two placements have the same number of LiDAR and similar InfraD in InfraLOB, the placement that two LiDARs on the diagonal give better uniformity and thus leads to better performance.
We also add two more sensors which increase the InfraD but contribute a little to InfraNUC.
This placement only slightly improves the accuracy, which shows the importance of point cloud uniformity, especially when the point cloud density is already large enough.
Scenario C is the two-way four-lane of urban road.
We place four LiDARs on each side of the road at the same location but set different pitch, and yaw angles to analyze how the rotation angles affect the accuracy.
We observe that setting appropriate rotation angles can achieve better performance, while too much downward tilt will lead to uneven point cloud distribution and damage the perception performance.
Note that we only change the pitch, yaw, and roll angle in scenario C. The pitch, yaw, and roll angle of LiDARs in scenarios A and B are all 0.
Through the above experiments, we find that InfraD and InfraNUC can be indicators of perception accuracy.
An optimal placement should produce point cloud data that have high density and good uniformity.

% the perception accuracy of the LiDAR placements is proportional to InfraD and inversely proportional to InfraNUC, the uniformity of the point cloud in the region.

% \subsection{Experiments for RLS Libaray}

% 2) Authenticity evaluation of point cloud generated by LiDAR simulation library

% In order to verify the authenticity of the proposed RLS library, we compare and analyze the point cloud generated by LiDAR from the RLS library, LiDAR from CARLA, and the real data set of the same LiDAR.
% For fair comparisons, we set the detection range, FOV range, and point frequency value of CARLA's ray cast LiDAR to be consistent with the LiDAR of the corresponding model in the LiDAR simulation library. We use the data collected by the LiDAR simulation library as the training set, use the open-source model framework to train the perception model, and finally test the method on the real LiDAR dataset to verify the matching degree between the simulation dataset and the real dataset. In this paper, we take Hesai pandar64, ouster os1-64, and Velodyne hdl-64 as examples to quantify the authenticity of the simulation library through experiments, and compare them with the CARLA ray cast LiDAR simulation data set under the same attribute conditions.

% \vspace{3mm}
\section{Conclusions}
In this paper, we investigate the correlation between infrastructure LiDAR placement and 3D detection performance using the simulate-and-evaluate pipeline.
To obtain a realistic simulation environment, we propose the RLS library that can simulate unique and realistic point cloud data.
For quick and traffic-irrelevant evaluation, we propose to use the InfraD and InfraNUC as surrogate metrics that characterize the density and uniformity of the point cloud data in the InfraLOB area.
Finally, we conducted experiments to evaluate infrastructure LiDAR placement in CARLA, validating the correlation between perception performance and surrogate metric of LiDAR configuration through representative 3D object detection algorithms.
Research in this paper sets a precedent for future work to optimize the placement of LiDARs precisely and quickly with realistic simulation.

\textbf{Acknowledgement.} 
This work is part of the OpenCDA Ecosystem~\cite{xu2023opencda} and is supported by the National Natural Science Foundation of China under Grant No. 62122010.
The research is supported by Shanghai Artificial Intelligence Laboratory, the National Key R\&D Program of China (Grant No. 2022ZD0160104) and the Science and Technology Commission of Shanghai Municipality (Grant No. 22DZ1100102).
\newpage

% \addtolength{\textheight}{-12cm}   % This command serves to balance the column lengths
                                  % on the last page of the document manually. It shortens
                                  % the textheight of the last page by a suitable amount.
                                  % This command does not take effect until the next page
                                  % so it should come on the page before the last. Make
                                  % sure that you do not shorten the textheight too much.

%%%%%%%%%%%%%%%%%%%%%%%%%%%%%%%%%%%%%%%%%%%%%%%%%%%%%%%%%%%%%%%%%%%%%%%%%%%%%%%%

%%%%%%%%%%%%%%%%%%%%%%%%%%%%%%%%%%%%%%%%%%%%%%%%%%%%%%%%%%%%%%%%%%%%%%%%%%%%%%%%

%%%%%%%%%%%%%%%%%%%%%%%%%%%%%%%%%%%%%%%%%%%%%%%%%%%%%%%%%%%%%%%%%%%%%%%%%%%%%%%%
% \section*{APPENDIX}

% Appendixes should appear before the acknowledgment.

% \section*{ACKNOWLEDGMENT}

% The preferred spelling of the word ÒacknowledgmentÓ in America is without an ÒeÓ after the ÒgÓ. Avoid the stilted expression, ÒOne of us (R. B. G.) thanks . . .Ó  Instead, try ÒR. B. G. thanksÓ. Put sponsor acknowledgments in the unnumbered footnote on the first page.

%%%%%%%%%%%%%%%%%%%%%%%%%%%%%%%%%%%%%%%%%%%%%%%%%%%%%%%%%%%%%%%%%%%%%%%%%%%%%%%%

% References are important to the reader; therefore, each citation must be complete and correct. If at all possible, references should be commonly available publications.

\bibliographystyle{IEEEtran}
\bibliography{IEEEexample}

\begin{thebibliography}{10}
\providecommand{\url}[1]{#1}
\csname url@rmstyle\endcsname
\providecommand{\newblock}{\relax}
\providecommand{\bibinfo}[2]{#2}
\providecommand\BIBentrySTDinterwordspacing{\spaceskip=0pt\relax}
\providecommand\BIBentryALTinterwordstretchfactor{4}
\providecommand\BIBentryALTinterwordspacing{\spaceskip=\fontdimen2\font plus
\BIBentryALTinterwordstretchfactor\fontdimen3\font minus
  \fontdimen4\font\relax}
\providecommand\BIBforeignlanguage[2]{{%
\expandafter\ifx\csname l@#1\endcsname\relax
\typeout{** WARNING: IEEEtran.bst: No hyphenation pattern has been}%
\typeout{** loaded for the language `#1'. Using the pattern for}%
\typeout{** the default language instead.}%
\else
\language=\csname l@#1\endcsname
\fi
#2}}

\bibitem{lang2019pointpillars}
A.~H. Lang, S.~Vora, H.~Caesar, L.~Zhou, J.~Yang, and O.~Beijbom,
  ``Pointpillars: Fast encoders for object detection from point clouds,'' in
  \emph{Proceedings of the IEEE/CVF conference on computer vision and pattern
  recognition}, 2019, pp. 12\,697--12\,705.

\bibitem{2018Complex}
M.~Simon, S.~Milz, K.~Amende, and H.~M. Gross, ``Complex-yolo: Real-time 3d
  object detection on point clouds,'' \emph{arXiv preprint arXiv.1803.06199},
  2018.

\bibitem{2017Vote3Deep}
M.~Engelcke, D.~Rao, D.~Z. Wang, C.~H. Tong, and I.~Posner, ``Vote3deep: Fast
  object detection in 3d point clouds using efficient convolutional neural
  networks,'' in \emph{2017 IEEE International Conference on Robotics and
  Automation (ICRA)}, 2017.

\bibitem{2018YOLO3D}
W.~Ali, S.~Abdelkarim, M.~Zahran, M.~Zidan, and A.~E. Sallab, ``Yolo3d:
  End-to-end real-time 3d oriented object bounding box detection from lidar
  point cloud,'' in \emph{ECCV 2018: "3D Reconstruction meets Semantics"
  workshop}, 2018.

\bibitem{2017PointNet}
C.~R. Qi, H.~Su, K.~Mo, and L.~J. Guibas, ``Pointnet: Deep learning on point
  sets for 3d classification and segmentation,'' in \emph{IEEE Conference on
  Computer Vision and Pattern Recognition (CVPR)}, 2017.

\bibitem{2014Beyond}
Y.~Xiang, R.~Mottaghi, and S.~Savarese, ``Beyond pascal: A benchmark for 3d
  object detection in the wild,'' in \emph{IEEE winter conference on
  applications of computer vision}, 2014.

\bibitem{2017VoxelNet}
Z.~Yin and O.~Tuzel, ``Voxelnet: End-to-end learning for point cloud based 3d
  object detection,'' in \emph{IEEE/CVF Conference on Computer Vision and
  Pattern Recognition (CVPR)}, 2017.

\bibitem{xu2022v2x}
R.~Xu, H.~Xiang, Z.~Tu, X.~Xia, M.-H. Yang, and J.~Ma, ``V2x-vit:
  Vehicle-to-everything cooperative perception with vision transformer,''
  \emph{arXiv preprint arXiv:2203.10638}, 2022.

\bibitem{bhover2017v2x}
S.~U. Bhover, A.~Tugashetti, and P.~Rashinkar, ``V2x communication protocol in
  vanet for co-operative intelligent transportation system,'' in \emph{2017
  International Conference on Innovative Mechanisms for Industry Applications
  (ICIMIA)}.\hskip 1em plus 0.5em minus 0.4em\relax IEEE, 2017, pp. 602--607.

\bibitem{olaverri2018connection}
C.~Olaverri-Monreal, J.~Errea-Moreno, A.~D{\'\i}az-{\'A}lvarez,
  C.~Biurrun-Quel, L.~Serrano-Arriezu, and M.~Kuba, ``Connection of the sumo
  microscopic traffic simulator and the unity 3d game engine to evaluate v2x
  communication-based systems,'' \emph{Sensors}, vol.~18, no.~12, p. 4399,
  2018.

\bibitem{chen2021pole}
G.~Chen, F.~Lu, Z.~Li, Y.~Liu, J.~Dong, J.~Zhao, J.~Yu, and A.~Knoll,
  ``Pole-curb fusion based robust and efficient autonomous vehicle localization
  system with branch-and-bound global optimization and local grid map method,''
  \emph{IEEE Transactions on Vehicular Technology}, vol.~70, no.~11, pp.
  11\,283--11\,294, 2021.

\bibitem{hu2022investigating}
H.~Hu, Z.~Liu, S.~Chitlangia, A.~Agnihotri, and D.~Zhao, ``Investigating the
  impact of multi-lidar placement on object detection for autonomous driving,''
  in \emph{Proceedings of the IEEE/CVF Conference on Computer Vision and
  Pattern Recognition}, 2022, pp. 2550--2559.

\bibitem{ma2021perception}
T.~Ma, Z.~Liu, and Y.~Li, ``Perception entropy: A metric for multiple sensors
  configuration evaluation and design,'' \emph{arXiv preprint
  arXiv:2104.06615}, 2021.

\bibitem{manivasagam2020LiDARsim}
S.~Manivasagam, S.~Wang, K.~Wong, W.~Zeng, M.~Sazanovich, S.~Tan, B.~Yang,
  W.-C. Ma, and R.~Urtasun, ``Lidarsim: Realistic lidar simulation by
  leveraging the real world,'' in \emph{Proceedings of the IEEE/CVF Conference
  on Computer Vision and Pattern Recognition}, 2020, pp. 11\,167--11\,176.

\bibitem{dosovitskiy2017carla}
A.~Dosovitskiy, G.~Ros, F.~Codevilla, A.~Lopez, and V.~Koltun, ``Carla: An open
  urban driving simulator,'' in \emph{Conference on robot learning}.\hskip 1em
  plus 0.5em minus 0.4em\relax PMLR, 2017, pp. 1--16.

\bibitem{caccia2019deep}
L.~Caccia, H.~Van~Hoof, A.~Courville, and J.~Pineau, ``Deep generative modeling
  of lidar data,'' in \emph{2019 IEEE/RSJ International Conference on
  Intelligent Robots and Systems (IROS)}.\hskip 1em plus 0.5em minus
  0.4em\relax IEEE, 2019, pp. 5034--5040.

\bibitem{hahner2019semantic}
M.~Hahner, D.~Dai, C.~Sakaridis, J.-N. Zaech, and L.~Van~Gool, ``Semantic
  understanding of foggy scenes with purely synthetic data,'' in \emph{2019
  IEEE Intelligent Transportation Systems Conference (ITSC)}.\hskip 1em plus
  0.5em minus 0.4em\relax IEEE, 2019, pp. 3675--3681.

\bibitem{hahner2021fog}
M.~Hahner, C.~Sakaridis, D.~Dai, and L.~Van~Gool, ``Fog simulation on real
  lidar point clouds for 3d object detection in adverse weather,'' in
  \emph{Proceedings of the IEEE/CVF International Conference on Computer
  Vision}, 2021, pp. 15\,283--15\,292.

\bibitem{hahner2022lidar}
M.~Hahner, C.~Sakaridis, M.~Bijelic, F.~Heide, F.~Yu, D.~Dai, and L.~Van~Gool,
  ``Lidar snowfall simulation for robust 3d object detection,'' in
  \emph{Proceedings of the IEEE/CVF Conference on Computer Vision and Pattern
  Recognition}, 2022, pp. 16\,364--16\,374.

\bibitem{gao2018object}
H.~Gao, B.~Cheng, J.~Wang, K.~Li, J.~Zhao, and D.~Li, ``Object classification
  using cnn-based fusion of vision and lidar in autonomous vehicle
  environment,'' \emph{IEEE Transactions on Industrial Informatics}, vol.~14,
  no.~9, pp. 4224--4231, 2018.

\bibitem{li2021learning}
Y.~Li, S.~Ren, P.~Wu, S.~Chen, C.~Feng, and W.~Zhang, ``Learning distilled
  collaboration graph for multi-agent perception,'' \emph{Advances in Neural
  Information Processing Systems}, vol.~34, pp. 29\,541--29\,552, 2021.

\bibitem{chen2017multi}
X.~Chen, H.~Ma, J.~Wan, B.~Li, and T.~Xia, ``Multi-view 3d object detection
  network for autonomous driving,'' in \emph{Proceedings of the IEEE conference
  on Computer Vision and Pattern Recognition}, 2017, pp. 1907--1915.

\bibitem{wang2020v2vnet}
T.-H. Wang, S.~Manivasagam, M.~Liang, B.~Yang, W.~Zeng, and R.~Urtasun,
  ``V2vnet: Vehicle-to-vehicle communication for joint perception and
  prediction,'' in \emph{European Conference on Computer Vision}.\hskip 1em
  plus 0.5em minus 0.4em\relax Springer, 2020, pp. 605--621.

\bibitem{xu2022opv2v}
R.~Xu, H.~Xiang, X.~Xia, X.~Han, J.~Li, and J.~Ma, ``Opv2v: An open benchmark
  dataset and fusion pipeline for perception with vehicle-to-vehicle
  communication,'' in \emph{2022 International Conference on Robotics and
  Automation (ICRA)}.\hskip 1em plus 0.5em minus 0.4em\relax IEEE, 2022, pp.
  2583--2589.

\bibitem{lei2022latency}
Z.~Lei, S.~Ren, Y.~Hu, W.~Zhang, and S.~Chen, ``Latency-aware collaborative
  perception,'' \emph{arXiv preprint arXiv:2207.08560}, 2022.

\bibitem{su2023uncertainty}
S.~Su, Y.~Li, S.~He, S.~Han, C.~Feng, C.~Ding, and F.~Miao, ``Uncertainty
  quantification of collaborative detection for self-driving,'' in \emph{2023
  IEEE International Conference on Robotics and Automation (ICRA)}.\hskip 1em
  plus 0.5em minus 0.4em\relax IEEE, 2023, pp. 5588--5594.

\bibitem{melotti2020multimodal}
G.~Melotti, C.~Premebida, and N.~Gon{\c{c}}alves, ``Multimodal deep-learning
  for object recognition combining camera and lidar data,'' in \emph{2020 IEEE
  International Conference on Autonomous Robot Systems and Competitions
  (ICARSC)}.\hskip 1em plus 0.5em minus 0.4em\relax IEEE, 2020, pp. 177--182.

\bibitem{fu2020depth}
C.~Fu, C.~Dong, C.~Mertz, and J.~M. Dolan, ``Depth completion via inductive
  fusion of planar lidar and monocular camera,'' in \emph{2020 IEEE/RSJ
  International Conference on Intelligent Robots and Systems (IROS)}.\hskip 1em
  plus 0.5em minus 0.4em\relax IEEE, 2020, pp. 10\,843--10\,848.

\bibitem{caltagirone2019lidar}
L.~Caltagirone, M.~Bellone, L.~Svensson, and M.~Wahde, ``Lidar--camera fusion
  for road detection using fully convolutional neural networks,''
  \emph{Robotics and Autonomous Systems}, vol. 111, pp. 125--131, 2019.

\bibitem{2012Car2X}
A.~Rauch, F.~Klanner, R.~Rasshofer, and K.~Dietmayer, ``Car2x-based perception
  in a high-level fusion architecture for cooperative perception systems,'' in
  \emph{Intelligent Vehicles Symposium}, 2012.

\bibitem{chen2019f}
Q.~Chen, X.~Ma, S.~Tang, J.~Guo, Q.~Yang, and S.~Fu, ``F-cooper: Feature based
  cooperative perception for autonomous vehicle edge computing system using 3d
  point clouds,'' in \emph{Proceedings of the 4th ACM/IEEE Symposium on Edge
  Computing}, 2019, pp. 88--100.

\bibitem{2020Learning}
N.~Vadivelu, M.~Ren, J.~Tu, J.~Wang, and R.~Urtasun, ``Learning to communicate
  and correct pose errors,'' 2020.

\bibitem{liu2019should}
Z.~Liu, M.~Arief, and D.~Zhao, ``Where should we place lidars on the autonomous
  vehicle?-an optimal design approach,'' in \emph{2019 International Conference
  on Robotics and Automation (ICRA)}.\hskip 1em plus 0.5em minus 0.4em\relax
  IEEE, 2019, pp. 2793--2799.

\bibitem{dybedal2017optimal}
J.~Dybedal and G.~Hovland, ``Optimal placement of 3d sensors considering range
  and field of view,'' in \emph{2017 IEEE International Conference on Advanced
  Intelligent Mechatronics (AIM)}.\hskip 1em plus 0.5em minus 0.4em\relax IEEE,
  2017, pp. 1588--1593.

\bibitem{kim2019placement}
T.-H. Kim and T.-H. Park, ``Placement optimization of multiple lidar sensors
  for autonomous vehicles,'' \emph{IEEE Transactions on Intelligent
  Transportation Systems}, vol.~21, no.~5, pp. 2139--2145, 2019.

\bibitem{kini2020sensor}
R.~R. Kini, ``Sensor position optimization for multiple lidars in autonomous
  vehicles,'' 2020.

\bibitem{li2019pu}
R.~Li, X.~Li, C.-W. Fu, D.~Cohen-Or, and P.-A. Heng, ``Pu-gan: a point cloud
  upsampling adversarial network,'' in \emph{Proceedings of the IEEE/CVF
  international conference on computer vision}, 2019, pp. 7203--7212.

\bibitem{xu2023opencda}
R.~Xu, H.~Xiang, X.~Han, X.~Xia, Z.~Meng, C.-J. Chen, C.~Correa-Jullian, and
  J.~Ma, ``The opencda open-source ecosystem for cooperative driving automation
  research,'' \emph{IEEE Transactions on Intelligent Vehicles}, vol.~8, no.~4,
  pp. 2698--2711, 2023.

\end{thebibliography}

\end{document}